\documentclass[10pt]{article}

\usepackage{afterpage}

\date{}
\author{}

\usepackage[linesnumbered, ruled, vlined]{algorithm2e}
\usepackage{etoolbox}
\AtBeginEnvironment{algorithm}{\linespread{.9}\selectfont}
\usepackage{multirow}
\usepackage{makecell}
\usepackage{mdframed}
\usepackage{float}
\usepackage[letterpaper]{geometry}
\usepackage{hicss}
\usepackage{times}
\usepackage[none]{hyphenat}
\usepackage{url}
\usepackage{mathtools}
\usepackage{soul}
\usepackage{gensymb}
\usepackage{cancel}
\usepackage{enumitem,kantlipsum}
\usepackage{latexsym}

\usepackage{lipsum}

\newcounter{num}

\usepackage{amsmath, amssymb, amsfonts}
\usepackage{textcomp}

\def\mat#1{\mathchoice{\mbox{$\displaystyle{\sf#1}$}}
{\mbox{$\textstyle\sf#1$}}
{\mbox{$\scriptstyle\sf#1$}}
{\mbox{$\scriptscriptstyle\sf#1$}}}
\def\m#1{\protect\mat #1}
\def\vec#1{\mathchoice{\mbox{$\displaystyle\bf #1$}}
{\mbox{$\textstyle\bf#1$}}
{\mbox{$\scriptstyle\bf#1$}}
{\mbox{$\scriptscriptstyle\bf#1$}}}
\def\v#1{\protect\vec #1}
\newcommand{\tr}{\mbox{$^{\top}$}}

\SetKwInput{KwInput}{Input}
\SetKwInput{KwOutput}{Output}

\begin{document}
\title{TSM: Measuring the Enticement of Honeyfiles with Natural Language Processing}

\author{Roelien C. Timmer \\
  UNSW Sydney \\
  Data61, Cyber Security CRC\\
  Sydney, Australia \\
  {\underline{r.timmer@unsw.edu.au}} \\\And
  David Liebowitz \\
  Penten\\
  UNSW Sydney\\
  Canberra, Australia \\
  {\underline{david.liebowitz@penten.com}}\\\And 
  Surya Nepal \\
  Data61\\
  Cyber Security CRC\\
  Sydney, Australia  \\
  {\underline{surya.nepal@data61.csiro.au}} \\\And 
  Salil S. Kanhere \\
  UNSW Sydney \\
  Cyber Security CRC\\
  Sydney, Australia\\
  {\underline{salil.kanhere@unsw.edu.au}} \\}

\maketitle
\begin{abstract} 
Honeyfile deployment is a useful breach detection method in cyber deception that can also inform defenders about the intent and interests of intruders and malicious insiders. A key property of a honeyfile, enticement, is the extent to which the file can attract an intruder to interact with it.  We introduce a novel metric, Topic Semantic Matching (TSM), which uses topic modelling to represent files in the repository and semantic matching in an embedding vector space to compare honeyfile text and topic words robustly. We also present a honeyfile corpus created with different Natural Language Processing (NLP) methods. Experiments show that TSM is effective in inter-corpus comparisons and is a promising tool to measure the enticement of honeyfiles. TSM is the first measure to use NLP techniques to quantify the enticement of honeyfile content that compares the essential topical content of local contexts to honeyfiles and is robust to paraphrasing.

\end{abstract}

\section{Introduction}

\subsection{Honeypots for Cyber Deception}

Cyber deception is increasingly being used in the defence of networks and information assets. A number of vendors are now advertising products that can detect, delay and mislead intruders and malicious insiders \cite{underwood2020}, generally with the creation and orchestration of honeypots. A honeypot is defined by Spitzner as {\em an information system resource whose value lies in unauthorized or illicit use of that resource} \cite{spitzner2003-1-honeypots}. 

Honeypots commonly mimic devices such as file servers (the popular Thinkst Canary, for example\footnote{see \url{https://canary.tools}}). They can be digital entities like document files, known as honeyfiles. A lot of valuable information is stored in documents such as financial reports, white papers, patent descriptions and contracts. Honeyfiles serve as digital traps just like other honeypots because they are fakes that mimic document files -- they have no
legitimate use, so any interaction with them is suspicious and suggests unauthorised access.
Honeyfiles can also provide information about the interests and intent of the adversary. If we know the content of a honeyfile, the fact that an adversary is choosing to search for, open or exfiltrate that content tells us what they are trying to discover or steal. As the number of documents stored in repositories and knowledge management systems grows, so does the benefit of using honeyfiles to protect them against unauthorised access. For a honeyfile to successfully detect unauthorised access, it must attract the attention of an intruder \cite{spitzner2003-1-honeypots}. The {\em enticement} of a honeyfile \cite{whitham2017automating} is a measure of how well it does this job.

{\color{black} Engaging the attention of an adversary to achieve a deception outcome requires some 
understanding of the adversary's goals and perceptions. Generalised models of deception developed by Bell~\cite{bell2003toward} and Whaley \cite{whaley1982toward}, based on observations from domains 
as diverse as the natural world, warfare and the practice of stage magic, describe identifying {\em channels} through which {\em ruses} can be communicated to the subject of the deception. In cyber deception, honeypots are typically 
``designed to seem as valuable as normal ones"~\cite{rauti2017survey} in order 
to entice the interaction that is essential to their use. Deception design can target the expected behaviour of an intruder, such as during the reconnaissance phase of a breach~\cite{fraunholz2018defending}. Some authors \cite{almeshekah2016cyber, wang2018cyber} 
advocate designing deceptions based on a thorough understanding of the 
adversary to exploit bias in their perception. 
We describe the limited literature on honeyfile enticement below, and argue for an approach to quantifying the characteristics of honeyfile content that suits modern document repositories so 
that we can entice the appropriate honeyfile interactions. A measure is important for the practical use of honeyfiles. 
It informs design decisions around text content, particularly in the era of generative deep learning models, because we must
ensure that honeyfiles are sufficiently enticing to operate effectively.}
 
\subsection{Honeyfiles} 
 
The first documented cyber defensive use of honeyfiles was Cliff Stoll's deployment of fake documents to trap an intruder on the Lawrence Berkeley National Laboratory networks in the 1980s \cite{stoll1988stalking, stoll2005cuckoo}. Stoll had been observing the intruder, and so could write honeyfiles with content and jargon to match their interests to entice the type and duration of interaction necessary to trace their location. Yuill {\em et al} \cite{yuill2004honeyfiles} described a system for intrusion detection in 
which a user places a honeyfile on a file server with a mechanism that would trigger an alert when the file is opened. The user would then know they had been compromised if they received an alert. This innovative early concept honeyfile system relied on manually created files, with content chosen to be enticing to hackers by virtue of file names signalling that it contains information like credentials or financial data. Bowen {\em et al} \cite{bowen2009baiting} developed the Decoy Document Distributor, a system to automatically generate honeyfiles using templates for documents like tax returns and bank statements. A number of honeyfile properties were defined to describe their design, deployment and deceptive effects, including notions of how believable, enticing and conspicuous the files are. These properties were defined in thought experiments that supposed probabilities of outcomes should an intruder be faced with a decision involving a honeyfile. In \cite{bowen2009baiting} and a series of related papers \cite{salem2011decoy, voris2012lost, voris2013bait, voris2015fox}, honeyfile use was explored with a number of experiments testing these properties. Of particular interest, Ben Salem {\em et al} \cite{salem2011decoy} investigated {\em enticingness} and {\em conspicuousness} in a study with 40 student subjects tasked with a scenario of theft from a desktop computer. They concluded that {\em conspicuousness comes first} for decoy effectiveness, that it is more important than enticingness in honeyfile access events.

\subsection{Motivation}

Whitham \cite{whitham2014design} developed a set of properties (similar to \cite{bowen2009baiting}) as design requirements for scalable, automated honeyfile generation, and followed up in \cite{whitham2017automating} a method to automatically create enticing honeyfile content with NLP techniques. However, a sophisticated enticement based on NLP techniques is lacking. In current literature, honeyfile enticement is measured by simple word count. The enticement of a document is quantified by counting all the words common to a honeyfile and files in its local directory, and those shared by the honeyfile and the rest of the file system. 

Two observations motivate our approach to enticement in this paper:
\begin{enumerate}[ itemsep=-1ex,leftmargin=0cm,itemindent=.5cm,labelwidth=\itemindent,labelsep=0cm,align=left,label=\textbf{\arabic*}),topsep=0pt] %
\item The experimental results obtained by Ben Salem {\em et al} \cite{salem2011decoy}, showing that the conspicuous placement of honeyfiles is more important than enticement, were derived from experiments on the local file system of a desktop computer. Users had to navigate the file system searching for files worth stealing, so placement had a significant impact. While the security of personal desktops remains a challenge, we believe that document theft is most damaging when it targets large scale repositories and knowledge management systems. Such systems store vast numbers of documents for governments and corporations, are indexed and searchable and present high-value targets for industrial espionage, state actors and other sophisticated adversaries. Searching for files in large repositories is likely to be through a search engine, not navigating a file system, making the text content of honeyfiles the primary means of promoting interaction with honeyfiles. 
\item Whitham's \cite{whitham2014design} enticement measure uses words common to real files and honeyfiles. This approach is limited by insensitivity to paraphrasing and the use of synonyms -- the same information expressed with different words with similar meaning would not count towards the enticement score.
\end{enumerate}

\subsection{Contributions}

We use NLP methods to address these limitations by proposing a novel measure we call Topical Semantic Matching (TSM) as follows. 
\begin{enumerate}[itemsep=-1ex,leftmargin=0cm,itemindent=.5cm,labelwidth=\itemindent,labelsep=0cm,align=left,label=\textbf{\arabic*}),topsep=0pt]
\item Topic modelling. We focus here on a scenario in which the adversary has access to a repository search engine and is looking for documents to steal. The indexed text content of the files is searched, so the enticement score must reflect how well honeyfile text represents the content of real files that may be a target. To generalise the idea of a local directory to subsets of a repository, we define the {\em local context} to be the set of documents files with a common theme or subject. We can define a local context in an indexed repository by the top $N$ results returned by a set of search terms. Topics are the words representing the key themes appearing in all the documents in a corpus, so we compare honeyfiles to real files by comparing honeyfile text to the topics representing the real file corpus. This approach can be used with a local file system and placing a honeyfile in a local directory. Topic modelling is described in Section~\ref{sec:topic_modelling}.
\item Semantic matching. The similarity between honeyfile and topics is quantified by the similarity of the {\em meanings} of the words. Semantic similarity compares words by their embeddings in a vector space where related words cluster together, and is described in Section~\ref{sec:semantic_matching}. 
\end{enumerate} 

We investigate TSM variants with the use of different topic modelling methods and aggregations of individual word similarities. Experiments on document corpora synthesised for this research show that TSM is a robust similarity measure when corpora derived from different category searches and domains are compared. This prepares the way to assess TSM as an enticement measure in an experimental setting.


Section~\ref{sec:NLP_methods} describes the NLP methods used and Section~\ref{sec:topical_semantic_matching} introduces TSM. Section~\ref{section:experiments} describes our honeyfile corpus and experimental results. Our paper finishes with limitations and a discussion in Section~\ref{sec:limitations} and a conclusion in Section~\ref{sec:conclusion}.

{\color{black}\section{Theory}\label{sec:NLP_methods}}
{\color{black}
The following section provides an overview of the four NLP methods used in this paper: topic modelling, semantic matching, doc2vec and text generation models. We limit the presentation to a brief summary and definition of the terminology that appear in subsequent sections, and refer the interested reader to accessible material for further information.}

\subsection{Topic Modelling}\label{sec:topic_modelling}
Topic modelling is a major subject of research in NLP that deals with extracting the main topics from texts. Topic models are particularly useful to extract hidden semantic structures. The output of topic models are clusters of similar words. The first topic models were created in the 1990s \cite{papadimitriou1998latent}. Topic models gained popularity when Blei, Ng, and Jordan introduced the Latent Dirichlet Allocation (LDA) model \cite{blei2003latent} in 2003. LDA is a hierarchical Bayesian model where every topic is modelled as a mixture of words, and a document is a mixture of topics. LDA has remained extremely popular, even as many other topic models have been published in recent years, such as the {\color{black} zero-shot cross-lingual \cite{bianchi2020cross}}, Stochastic Block Model \cite{gerlach2018network}, lda2vec \cite{moody2016mixing} and embedded \cite{dieng2020topic} topic models. {\color{black}A readable tutorial on LDA can be found in Darling\cite{darling2011theoretical}}.

In our comparative analysis, we use LDA \cite{blei2003latent} and SBM \cite{gerlach2018network} topic models. We selected LDA because it is the most popular, and SBM as it has emerged recently with claims to better the performance of LDA.

\subsection{Semantic Matching}\label{sec:semantic_matching}
Semantic matching, as used below, compares the meanings of words. Quantifying the semantic relationship between words uses vector representations that embed them in a high dimensional space -- a semantic vector space. These embeddings can be created in a number of ways using words and their contexts. The context of a word is the words that appear just before and after it. 

Early representations~ \cite{turney2010vector} used matrices to represent the co-occurrence of words and contexts, counting the number of times each word appears within each context. The rows and columns of these matrices represent the words and contexts as large, sparse vectors, or lower-dimensional, dense vectors after matrix factorisation~ \cite{goldberg15lesssons}. Neural embeddings have emerged more 
recently, notably word2vec~ \cite{mikolov2013efficient, mikolov2013distributed}, which computes embeddings using a neural network that predicts words or contexts (it has versions that do both). Similar embeddings can be derived from other approaches like 
GloVe~ \cite{pennington2014glove} or FastText~ \cite{bojanowski2017enriching}. 

The vector representations we use to compute enticement scores do not depend on any embedding method, provided all the words are in the same semantic vector space. We compare the vector similarity of two words using the normalised inner product of the vectors. 
Let $\v x$ and and $\v y$ be (column) vectors of length $f$ representing words $u$ and $v$ respectively:
\begin{equation*}
\v x  =  (x_1, x_2, ..., x_f) {\rm \ and \ }
\v y  =  (y_1, y_2, ..., y_f). 
\end{equation*}
The inner product of $x$ and $y$ is the cosine similarity, and quantifies how similar the words $u$ and $v$ are in meaning:
\begin{equation}
\label{eq:cosine_similarity}
\cos(\theta_{uv}) = \frac{x \cdot y}{ \lVert x \rVert \lVert y \rVert}
 =\frac{\sum_{i=1}^{f} x_i y_i}{\sqrt{\sum_{i=1}^{f} x_i^2} \sqrt{\sum_{i=1}^{f} y_i^2}}
\end{equation}
with $\|x\|$ and $\|y\|$ being the Euclidean norm which is the vector length. The cosine similarity can range from -1 to 1. When words are more similar, their vectors are closer to each other and the cosine similarity is closer to 1.

For ease of reference, we normalise the cosine similarity to the range $[0, 1]$ to yield a semantic similarity between words $u$ and $v$:
\begin{equation}
\label{eq:cosine_similarity_2}
S_{uv} = \frac{\cos(\theta_{uv}) + 1}{2}
\end{equation}

\begin{figure}[b] 
\centerline{\includegraphics[width=0.8\linewidth]{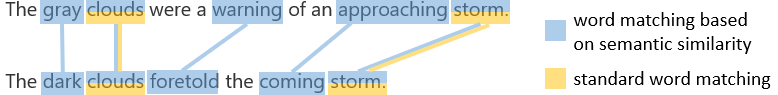}}
\caption{The difference between word matching based on semantic similarity and the common words.}
\label{fig:semantic_similarity}
\end{figure}

Fig. \ref{fig:semantic_similarity} visualises why comparing words based on their semantic similarity is useful. The second sentence is a paraphrased version of the first. 
Standard word matching only detect two common words matches while semantic matching detects five.

\subsection{Doc2vec}\label{sec:doc2vec}
Sentences, paragraphs and whole documents can be represented by embeddings in a vector space. The embeddings represent the semantic content of the texts they represent, so that similarity measures may be applied. 
Doc2vec is a popular approach by Mikolov and Le \cite{le2014distributed}, and uses a similar methodology to word2vec. {\color{black} For details on embedded representations and their use in NLP, see the tutorial by Goldberg \cite{goldberg2016primer}.} A reasonable approach to enticement with semantic matching would be to compare doc2vec embeddings. To our knowledge, this has not been proposed in the literature, and we test it experimentally below. 
There are two different implementations of Doc2vec, the Distributed Memory Model of Paragraph Vectors (PV-DM) and the Distributed Bag-of-Words version of Paragraph Vector (PV-DBOW). 
Mikolov and Le showed based on experiments that PV-MD performs best on average \cite{le2014distributed}. In this paper, we, therefore, use the PV-DBOW implementation of Doc2vec. In our experiments , we test whether doc2vec embeddings can be used as the basis for a simple enticement measure. The results show that the TSM is a better enticement measure, probably because the doc2vec embeddings are strongly influenced by language other than the key topics in the documents. 

\subsection{Honeyfile Text Content}\label{sec:honeyfile_content}
The text content of honeyfiles can be created manually, but this is time-consuming and prevents deployment at scale. Bowen {\em et al}~ \cite{bowen2009baiting} automated honeyfile creation with templates based on common documents. Whitham~ \cite{whitham2017automating} used the text in files in a target directory to populate a honeyfile. This approach starts with Part-of-Speech (POS) tagging of n-grams in the reference text to assemble text fragments. Then a file from the target directory is used as a template by substituting text fragments in place, sampling from the assembled fragments while trying to match the POS-tagging of the n-gram fragments in the template. 

Recent developments in language models have given rise to enabling technology that can mimic human text remarkably well. Such automatic text generation received lots of attention in 2018 when the Generative Pre-trained Transformer, better known as GPT, was released by OpenAI~ \cite{radford2018improving}. An improved model, GPT-2~ \cite{radford2019language}, was released soon after in 2019. In 2020 GPT-3 followed~ \cite{brown2020language}. The number of parameters in the model increased dramatically over these three different versions, from 117 million parameters to 1.5 billion parameters and eventually to 175 billion parameters. There are now specialised text generation models, for example, a model that generates coherent and cohesive long-form texts~ \cite{cho2018towards}. Recent publications have focused specifically on fake text or fake code generation for honeyfiles~ \cite{karuna2018enhancing, karuna2018generating, nguyen2021honeycode}. 
A cyber security company [anon] has a product that creates and manages the lifecycle of honeyfiles to protect document repositories [anon]. We use honeyfiles generated by this technology to test the proposed measure, as described in Section~\ref{section:experiments}. 
This technology supports training the text generation model with three approaches: a POS-tagged substitution model, a similar model that uses dependency parsed tokens (DPT) instead of POS-tagged n-grams, and a GPT-2 model that can be fine-tuned on a corpus. It supports using the standard Lorem Ipsum dummy text instead of a trained model. Lorem Ipsum is a pseudo-Latin text often used as a placeholder~\cite{team2019people}.

The Lorem Ipsum text should not match any local context. Because 
the POS-tagged and DPT train a 
text substitution model using fragments from the local context, the honeyfile contains real text, but in a garbled form. 
GPT-2 mode uses a model fine-tuned on the local context corpus. It contains text semantically similar to the local context due to fine-tuning and is usually more realistic than the other methods. 

\section{Topical Semantic Matching}
\label{sec:topical_semantic_matching}
We propose the Topical Semantic Matching (TSM) enticement score for honeyfiles. This novel enticement measure is based on topic modelling to capture the topics contained in the local context and semantic matching to measure the similarity between the honeyfile and topics.
TSM can be calculated with a topic model of choice. 

We refer to the honeyfile text as $h$ and the local context text as $l$. The honeyfile $h$ is a set of $N_K$ words:
\begin{equation}
h = \{ u_1, u_2, \ldots , u_{N_K}\}
\end{equation}

The local context $l$ is composed of $M$ words from all the files in the local context:
\begin{equation}
l = \{w_1, w_2, \ldots, w_M\}
\end{equation}

We preprocess the honeyfile $h$ and local context $l$ by removing stop words, applying named entity recognition (NER) and lemmatisation. These three steps are all common NLP preprocessing practices, giving
$h'$ and $l'$, with $N_{K'}$ and $N_{M'}$ words respectively. 
Then we extract the topics from $L'$ in the form of a set of $N_T$ topic words to give $t = \{ v_1, v_2, \ldots, v_{N_T} \}$.

To extract the topics, a topic model of choice can be used. For our comparative analysis we selected LDA because of its popularity, and SBM as it has emerged recently with claims to better performance than LDA.

Next, we want to quantify the similarity in meaning between the honeyfile and the local context as represented by the topic words. To do this, we construct matrices $\m H $ and $\m T$ from sets $h'$ and $t$ by representing each word in $h'$ and $t$ by their embedding vectors:
\begin{equation*}
\m H  =  \left[ \v x_1, \ldots, \v x_{N_K'} \right] {\rm \ and \ }
\m T  =  \left[ \v y_1, \ldots, \v y_{N_T} \right]
\end{equation*}

where we use $\v x$ for honeyfile word vectors and $\v y$ for topic word vectors.

This representation has computational advantages when computing similarity scores by using efficient linear algebra implementations available in most numerical libraries \cite{fatahalian2004understanding}. To do this we normalise the columns of $\m H$ and $\m T$ to length 1, calculate the matrix product $\m H \m T\tr$ and scale each element to the range $[0, 1]$ using Equation \ref{eq:cosine_similarity_2} to give 
$\m S_{HT}$, the semantic similarity between all word pairs from $h'$ and $t$.

\begin{figure*}[htbp]
\centerline{\includegraphics[width=.6\linewidth]{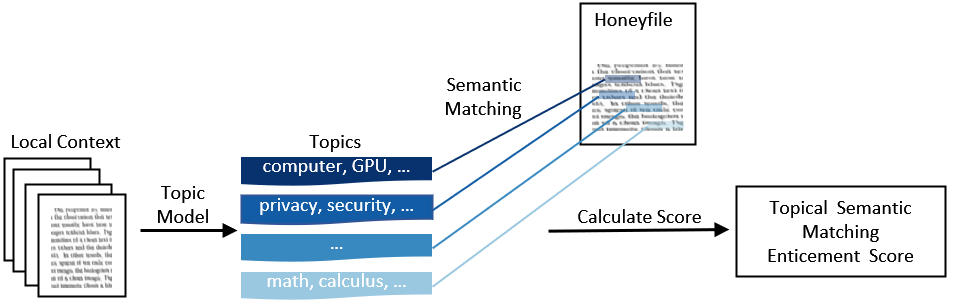}}
\caption{To calculate TSM, a topic model extracts the topics from the local context. TSM is the weighted average of the semantic similarities between these topics and the honeyfile words above a certain threshold.}
\label{fig:mesh1}
\end{figure*}

We explore two methods to aggregate these similarities into a score: averaging all similarity scores between the topics and the honeyfiles words, and thresholding before averaging. As data thieves often search for specific keywords, we expect that a high threshold is most suitable to measure honeyfile enticement.
This gives the final TSM enticement score $E$ of the honeyfile $h$ with respect to the local context. Fig. \ref{fig:mesh1} shows a visual representation of the TSM measure and Algorithm \ref{alg:topic} outlines the pseudo-code \footnote{The implementation of TSM can be found on \url{https://
github.com/RoelTim/tsm-honeyfile-nlp-enticement}.}. 

\begin{algorithm}[t]
\small
\DontPrintSemicolon
\KwInput{honeyfile $h$, local context $l$, threshold $\delta$}
\KwOutput{enticement score $E$}
$l'$, $h'$ : preprocess $l$ and $h$ by removing stop words, applying lemmatisation and named entity recognition

$t$ : extract the topics from $l'$

$\m L, \m H$ : represent each word of $l'$ and $t$ by their embedding

$\m S_{\m H\m T}$ : normalise all columns of $\m H$ and $\m T$ to length 1, calculate the matrix product $\m H\m T^T$and scale to the range[0, 1] using Equation \ref{eq:cosine_similarity_2}. 

\If{thresholding method}
{
$E = \sum s_i / (N_{K'}*N_T)$ for $s_i \in \m S_{\m H\m T}, s_i \geq \delta$, 
$0\leq \delta \leq 1$
}
\Else{use the averaging method\\
$E = \sum(\m S_{\m H\m T})/(N_{K'}*N_T)$
}
\caption{\textbf{TSM Score}}
\label{alg:topic}
\end{algorithm}

\section{Comparative Analysis} \label{section:experiments}
In this section, we explain how we created our honeyfile corpus and we show that the TSM measure with threshold performs best.

\subsection{Honeyfile and Local Context Generation}

\begin{figure}[htbp] 
\centerline{\includegraphics[width=.8\linewidth]{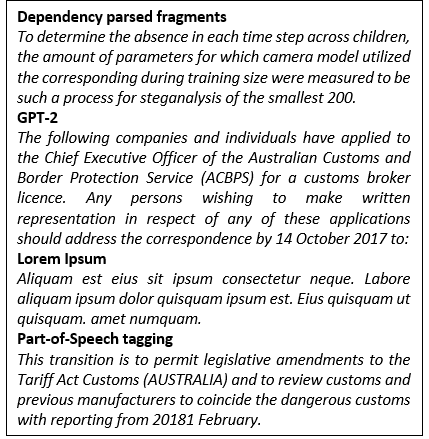}}
\caption{Text snippets from the honeyfiles.}
\label{fig:honeyfile_textsnippet}
\end{figure}

We could not find a suitable corpus that contains honeyfiles and local context files for our experiments. We thus generated one ourselves. We scraped files from the Internet that serve as our local context and as input for the generation of the honeyfiles. We selected technical files as hackers are often interested in intellectual property and other technical documents. For this paper, the technical files are academic papers from different disciplines and a set of official customs notices published by the Australian Government.

The experimental corpora contain four different types of files of which three are academic papers with author keywords\footnote{These papers are downloaded from Web of Science. We randomly selected papers with the corresponding author keyword and English as the main language. The author keyword `plants' refers to biological plants and not to industrial plants.} theater, computer architecture and plants. The fourth type of file is  Australian customs notices\footnote{Australian Custom Notices are online at \url{https://www.abf.gov.au/help-and-support/notices/australian-customs-notices}. We downloaded files in the date range 1996-01 to 2020-39.}.

\begin{table}[t]
\footnotesize
\centering
\caption{\label{font-table}The number of web scraped local files also serve as the training corpus for the honeyfiles.\vskip 3pt}
\label{tab:local-context-table}
\begin{tabular}{|l|l|}
\hline
\makecell[l]{Category Local Files}& \makecell[l]{Number} \\
\hline
Australian customs notices & 1460\\
Papers about `theater' & 100 \\
Papers about `computer architecture' & 100\\
Papers about `plants' & 140 \\ 
\hline
Total & 1800 \\
\hline
\multicolumn{2}{l}{\makecell[l]{$^{\mathrm{a}}$Every local contexts consist of 5, 10 or 20\\files of a category.}}
\end{tabular}
\end{table}

\begin{table*}[htbp]
\footnotesize
\caption{Number of Honeyfiles Generated with Different Generation Methods and Based on Different Corpora.}
\begin{center}
\begin{tabular}{|l|l|l|l|l|l|}
\hline
 & \multicolumn{4}{c|}{Honeyfile Training Corpus}&\\\hline
 & \makecell[l]{Australian custom notices} & Theater & \makecell[l]{Computer architecture} & Plants & Total \\\hline
 GPT-2& 103 & 25 & 25 & 25 & 178 \\\hline
\makecell[l]{Lorem Ipsum} & N/A & N/A & N/A & N/A & 160 \\ \hline
\makecell[l]{POS-tagging}& 100 & 20 & 20 & 20 &160 \\ \hline
\makecell[l]{Dependency Parsed Tokens}& 100 & 20 & 20 & 20 &160 \\ \hline
Total& 303 & 65 & 65 & 65 & 498 \\ \hline
\end{tabular}
\end{center}
\end{table*}

Table \ref{tab:local-context-table} shows the number of files per category. In this paper, a local context consists of 5, 10 or 20 different local context files from the same category. For the customs notices, we selected files for a local context based on release dates. For example, for a local context of size 10, the first file in the custom local context is 1996-01 and the last is 1996-10.

Each category of local context files also serves as the input to generate honeyfiles. Four honeyfile generation methods are used:  GPT-2 based, POS-tagging, DPT or replacement with Lorem Ipsum text. For all the four methods we started with a template\footnote{We have eight different templates which are derived from the local context files. The eight templates are: 1996-01, 1996-17, 2018-02, 2019-40, 2020-39, 08552374, 0021989420918654 and wild-useful-herbs-of-aktobe-region-western-kazakhstan.}. The honeyfiles generated mimic the layout of these templates.

The GPT-2 based honeyfiles were generated by fine-tuning the pre-trained GPT-2 medium model on the four different local context corpora. For example, a `computer architecture' honeyfile was generated by fine-tuning on the `computer architecture' corpus. The POS-tagging honeyfiles were generated by replacing the words in the template with another word that has the same POS-tagging. The DPT files were generated by imitating the DPT structure of the template. The Lorem Ipsum honeyfiles were generated by replacing the text in the templates with Lorem Ipsum. Fig. \ref{fig:honeyfile_textsnippet} shows text snippets of the different types of honeyfiles generated.

We expect that a honeyfile $h$ generated based on a corpus has a high enticement score when compared to a local context file from that corpus. For example, a honeyfile that was trained on `theater' files should have a high enticement score if it is compared with a local context from the `theater' category. 

\subsection{Selecting the Best Enticement Measure}

In this section, we show that TSM with thresholding is best in measuring enticement and show the results on the corpus we created. 

\begin{figure*}[t] 
 \centering
 \includegraphics[width=0.8\linewidth]{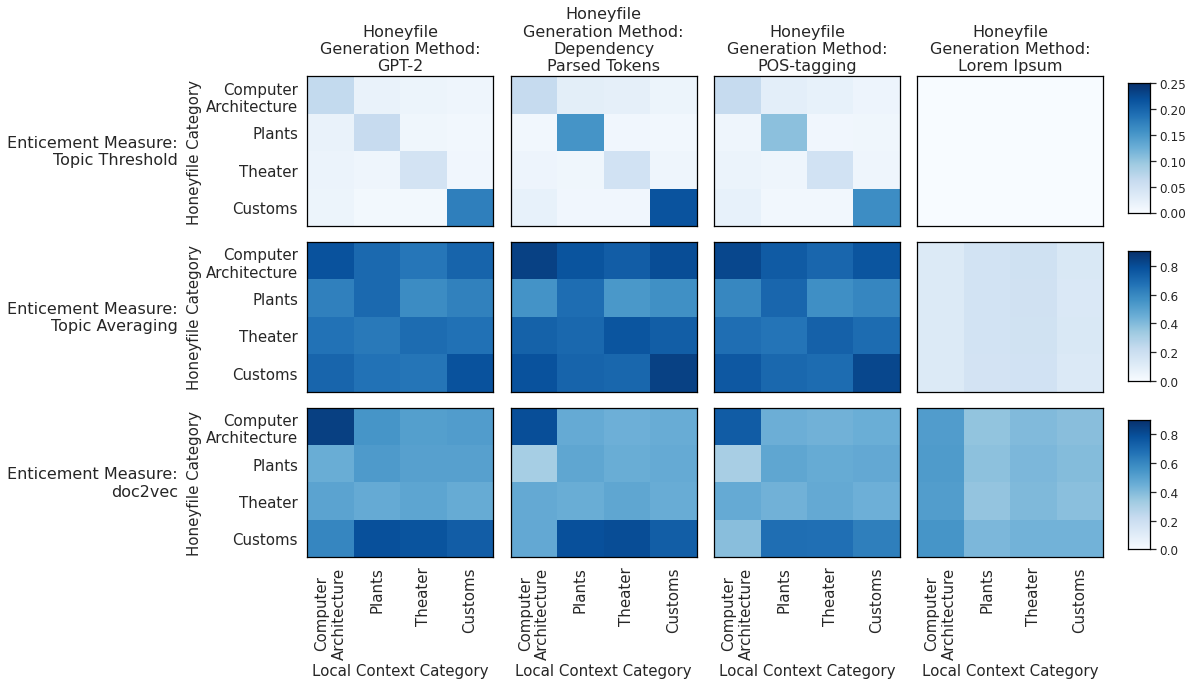}
 \caption{The 50th percentile of the enticement scores between the honeyfiles matched with local context files. }
 \label{fig:heatmap}
\end{figure*}

\begin{figure*}[h]
 \centering
 \includegraphics[width=0.8\linewidth]{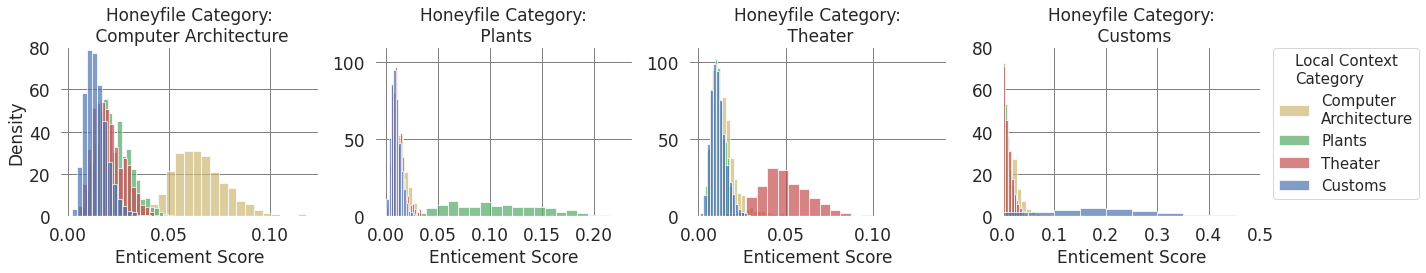}
 \caption{The distribution of the TSM enticement score.}
 \label{fig:distributions}
\end{figure*}

Fig. \ref{fig:heatmap} shows that TSM with thresholding has a higher enticement score for honeyfiles matched with local contexts from the same category. This heatmap shows that the doc2vec and TSM without thresholding perform poorly. Intuitively this makes sense as TSM with thresholding only takes into account words that have a high semantic similarity with the main topics. TSM with thresholding filters out noise as it ignores words that have a low semantic similarity. The three diagonals on the top left corner show that the enticement scores is higher between honeyfiles and local contexts that correspond to the same corpus. For example, a honeyfile that was generated based on the `computer architecture' corpus matched with a local context based on the `computer architecture' corpus has a relatively high enticement score. The topic model used for the TSM scores is LDA with five topics each consisting of ten words. Each local context consists of 10 files. Fig. \ref{fig:distributions} shows the distribution of the enticement scores of TSM with a threshold of 0.9 and 10 files per local context.

We include doc2vec in the comparison as doc2vec takes into account all words of a document \cite{le2014distributed}, while TSM only extracts the main topics. We trained the doc2vec model on the preprocessed Australian custom notices and the academic articles. Next, we trained our doc2vec model on a data set similar to the local context $l$. The doc2vec model can use the Distributed Memory Model of Paragraph Vectors (PV-DM) or the Distributed Bag of Words version of Paragraph Vector (PV-DBOW)\footnote{Our parameters are max\_epochs = 100, vec\_size = 20, alpha = 0.025 and embedding model = Distributed Memory (PV-DM).}. The first (i.e., PV-DM) generally gives better results, but requires a longer training time. After training the doc2vec model we extract the embedding of the honeyfile $h$ and the local context $l$. As a final step, to calculate the enticement score, we calculate the cosine similarity between these two vectors. 

Fig. \ref{fig:heatmap} shows that for the doc2vec measure the enticement scores are not necessarily higher between honeyfiles and local contexts that correspond to the same corpus. As expected, it is better to focus on the main topics than on the whole document as with doc2vec. 

All these results are consistent over the three main honeyfile generation methods which are based on GPT-2, DPT and POS-tagging. The honeyfiles containing Lorem Ipsum text have enticement scores of or close to zero. This is in line with our expectations, as Lorem Ipsum text is not related to any of the corpora.

The next experiments use default variables unless stated otherwise. The local context size is 10 and the topic model is LDA\footnote{We use the GENSIM LDA model: \url{https://radimrehurek.com/gensim/models/ldamodel.html}}. For LDA we select five topics each consisting of ten words. We aggregate the GPT-2, DPT and the POS-tagging honeyfiles. We leave out the Lorem Ipsum honeyfiles as we consider them weak honeyfiles. All the heatmaps show the 50th percentile of the enticement scores.

\subsection{Experimenting with Thresholds}
\begin{figure}[htbp]
 \centering
 \includegraphics[width=0.8\linewidth]{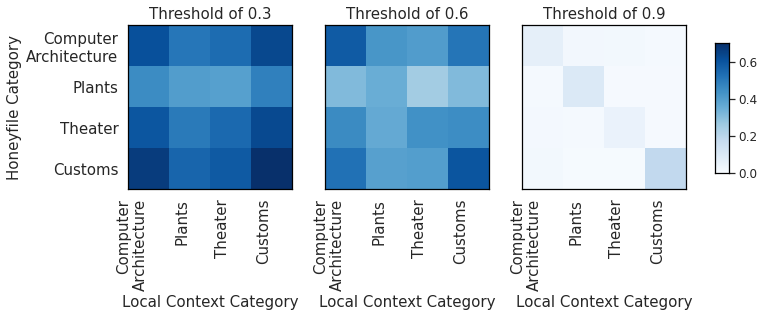}
 \caption{
TSM works best with a high threshold.} 
 \label{fig:thresholds}
\end{figure}
\begin{figure}[htbp]
\centering
\includegraphics[width=0.8\linewidth]{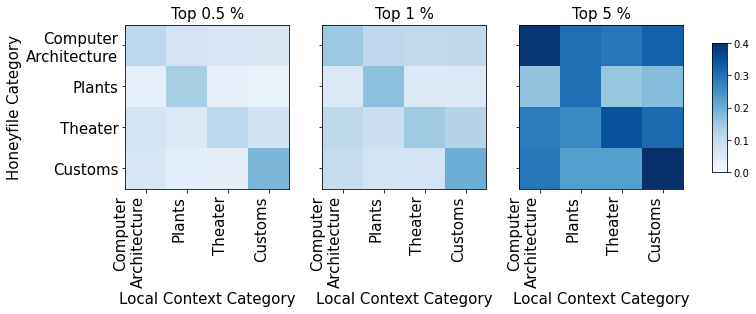}
\caption{Aggregating the top 0.5\% semantic similarities give similar results to TSM with threshold.}
\label{fig:percentage}
\end{figure}

We experimented with several TSM similarity thresholds. Fig. \ref{fig:thresholds} shows that a high threshold of 0.9 yields a good result. This is not surprising as only very similar words get matched with a high semantic similarity. This means that irrelevant words are ignored, and that enticement is well represented by distances between the most similar topics and honeyfile words. 
Instead of aggregating the semantic similarities that surpass a certain threshold, we can aggregate only the highest similarities. Fig. \ref{fig:percentage} shows that selecting the top 0.5\% of the highest semantic similarities leads to good results. Intuitively this makes sense, as in line with the thresholding method, we only capture the topics that have a high similarity with the text in the honeyfile.

\subsection{Experimenting with Topics Models}
\begin{figure}[htbp]
 \centering
 \includegraphics[width=0.8\linewidth]{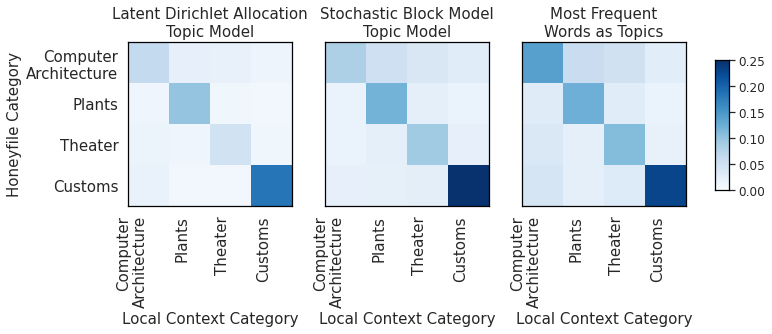}
 \caption{
 TSM score are comparable for different topic models and when the top 50 words are used.}
 \label{fig:scores_topic_models}
\end{figure}

TSM does not depend on any particular topic model. Fig. \ref{fig:scores_topic_models} shows that the results of the LDA and SBM topic models are comparable. For the SBM topic model we select the default level $l=1$ and the number of words $n=50$\footnote{Link to the code of the SBM topic model: \url{https://github.com/martingerlach/hSBM_Topicmodel}}. Next, we select the 50 words that contribute most to the $l=1$ topics.

We experimented with replacing the topic models by selecting the 50 most common words of the preprocessed local context files. Fig. \ref{fig:scores_topic_models} shows that the results of selecting the 50 most common words are similar to applying a topic model.

The biggest advantage of selecting the most common words is the low computational cost. Selecting the most common words is faster than running a topic model, although in general better results can be expected using topic modelling.

\subsection{Experimenting with Local Context Sizes}

\begin{figure}[htbp]
 \centering
 \includegraphics[width=0.8\linewidth]{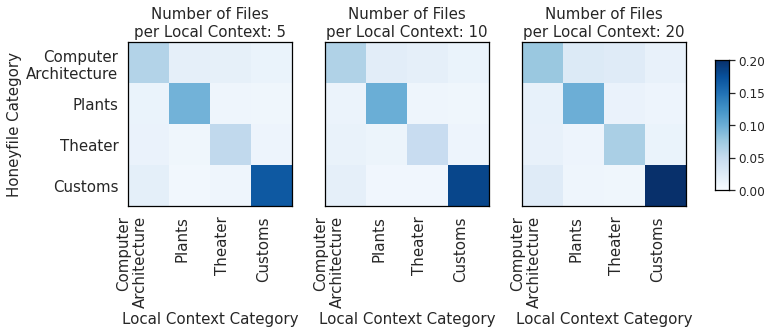}
 \caption{The TSM scores are comparable when the local contexts differ in size.}
 \label{fig:local_cont_sizes}
\end{figure}

In practice, the local contexts vary in size. Therefore, we experimented with different local context sizes of 5, 10 and 20 files. Fig. \ref{fig:local_cont_sizes} shows that the performance of TSM is similar when the local context size varies. The heatmaps shows that TSM results are almost identical for the different local context sizes.

{\color{black}
\section{Limitations and Discussion} \label{sec:limitations}
TSM is proposed as a metric to measure the enticement of the text content in a honeyfile. It is aimed at a context in which honeyfiles protect real documents in a repository accessible by search. We argue 
that the honeyfile text should be semantically similar to the topics represented in the real files so 
that they are as likely as the real documents to be returned by meaningful search terms. This is necessary for the honeyfiles
to work, since interaction is the signal that drives the defensive benefits. 

Enticement as a measure should reflect how well the honeyfile content attracts the attention of intruders. We show 
experimentally that TSM correctly reflects the semantic similarity of honeyfiles and the topics of text corpora. This suggests that it is a plausible candidate for an effective enticement measure. While there is a growing body of experimental work in cyber deception, such as the Tularosa Study~\cite{ferguson2018tularosa, ferguson2021decoy}, it does not address the specific interactions associated with honeyfiles or text content. We intend to test the effectiveness of TSM as an enticement metric in human trials with text-based interactions. 

TSM, as implemented here, cannot distinguish between homonyms. For example, the word `good' is used in Australian Customs notices to mean a product while in other documents it is used as an adjective. It could potentially be improved using part-of-speech or dependency parsed word embeddings \cite{suleiman2019using, levy2014dependency}. 

A limitation likely to be seen in practice with small corpora is the constrained ability of topic models to extract reliable topic word distributions. Approaches such as topic cropping \cite{tran2013topic} might alleviate this somewhat, but small training corpora will be challenging from a text generation perspective as well. 

A general limitation of honeyfile research is the relatively small literature and absence of standard data sets. Thus, there is no golden data set that we can test our TSM measure on. We anticipate that our future user study will provide a data set that can be used for this purpose.

\section{Conclusion} \label{sec:conclusion}

In this paper, we used an NLP-based approach we call Topic Semantic Matching to develop a measure of the enticement of honeyfiles. TSM compares words in the honeyfiles to topics representing the real documents they protect. We show experimentally that TSM 
with a high threshold performs well at comparing the semantic content of honeyfiles to the corpora that its generative models trained on
relative to comparable corpora. Similar results are achieved with alternative topic models.

The key advantage of TSM is that it is robust to paraphrasing and the use of synonyms through the use of semantic matching. The existing common word count measure accounts for words that are exactly the same, while TSM accounts for similar meanings of words.

TSM is well suited to contexts in which real documents and honeyfiles are stored in a repository accessed via search terms, and we 
believe it is a promising candidate to evaluate as a practical enticement measure. Currently, we are working on a study that tests the TSM measure and other metrics on human subjects. We plan to also investigate quantitative measures of other honeyfile properties, such as realism and the presence of sensitive information. Realism, presence of sensitive information and enticement are often in conflict with each other. For example, a honeyfile that is almost a duplicate of a document that we want to protect is considered highly enticing and realistic but is also more likely to contain sensitive information.  
}

\section*{Acknowledgment}

The authors acknowledge the support of the Commonwealth of Australia, the Cybersecurity Cooperative Research Centre Limited, and Penten for supporting this work. In particular, we would like to thank Michael Longland, Lachlan Henderson and Nicholas Mobbs of Penten for their help in creating the honeyfile test corpus.

\bibliographystyle{IEEEtran}
\bibliography{deception}

\end{document}